\documentclass[10pt, a4paper]{article}

\usepackage[final]{lrec2026} 
\usepackage{booktabs}
\usepackage{subcaption}
\usepackage{amsmath}

\title{Hindsight Quality Prediction Experiments in Multi-Candidate Human-Post-Edited Machine Translation}

\name{Malik Marmonier \quad Benoît Sagot \quad Rachel Bawden} 

\address{Inria, Paris Center\\
         \{firstname.lastname\}@inria.fr\\}

\abstract{
This paper investigates two complementary paradigms for predicting machine translation (MT) quality: source-side difficulty prediction and candidate-side quality estimation (QE). The rapid adoption of Large Language Models (LLMs) into MT workflows is reshaping the research landscape, yet its impact on established quality prediction paradigms remains underexplored. We study this issue through a series of ``hindsight'' experiments on a unique, multi-candidate dataset resulting from a genuine MT post-editing (MTPE) project. The dataset consists of over 6,000 English source segments with nine translation hypotheses from a diverse set of traditional neural MT systems and advanced LLMs, all evaluated against a single, final human post-edited reference. Using Kendall's rank correlation, we assess the predictive power of source-side difficulty metrics, candidate-side QE models and position heuristics against two gold-standard scores: TER (as a proxy for post-editing effort) and COMET (as a proxy for human judgment). Our analysis yields three primary findings: (1) On the source side, the predictive power of difficulty metrics is highly contingent on the reference metric used; features that strongly correlate with COMET (e.g., segment length, neural predictors) show much weaker correlation to TER. (2) On the candidate side, we find a significant mismatch between QE model rankings and final human-adjudicated quality, and further show that modern QE metrics are significantly more aligned with the quality of traditional neural MT outputs than with those from general-purpose LLMs. (3) While we confirm a statistically significant positional bias in document-level LLMs (i.e., the tendency for translation quality to degrade for segments occurring later in a document) its practical impact on translation quality appears to be negligible. These findings highlight that the architectural shift towards LLMs alters the reliability of established quality prediction methods while simultaneously mitigating previous challenges in document-level translation.
 \\ \newline  }

\begin{document}

\maketitleabstract

\section{Introduction}
The ability to anticipate machine translation (MT) quality is valuable to both researchers and professional translators. It is typically addressed through two complementary paradigms, which we group under the umbrella term ``quality prediction.'' 

On the source side, translation difficulty prediction aims to estimate translation effort a priori by analyzing the source text. This can inform decisions on resource allocation and help avoid wasted effort on segments unlikely to meet minimal quality requirements for human post-editing \citep{fernicola-etal-2023-return}. Conversely, this can also drive the selection of more challenging source segments for the creation of evaluation sets in MT research \citep{proietti2025estimating}. On the candidate side, Quality Estimation (QE; cf.~\citealt{callison-burch-etal-2012-findings}) assesses the quality of a machine-generated translation without a target reference, enabling downstream tasks such as triaging translations by degrees of post-editing required, or selecting the best hypothesis among multiple candidates. 

These approaches were largely developed and benchmarked within an MT landscape that has been transformed in complex ways due to the rapid rise of large language models (LLMs) and their widespread adoption in translation use cases \citep{chatterji2025how}. LLMs, thanks to their general-purpose nature, have spurred massive industrial investment far exceeding that of specialized MT systems. A notable result of this scaling is a dramatic expansion of model capabilities with regard to context-window length, making document-level translation widely accessible for the first time. Yet, the practical consequences of these evolutions for the established methods of quality prediction remain underexplored.

This paper investigates this issue through a series of ``hindsight'' experiments conducted on a unique dataset with substantial ecological validity, as it is the byproduct of a genuine, multi-candidate post-editing project. For each of over 6,000 source segments, this dataset contains nine translation hypotheses from a diverse set of systems---including traditional NMT models and advanced LLMs---operating at various ``granularities'' (at the sentence or the document level). Because the gold-standard reference is the result of post-editing, the TER metric \citep{olive2005global, snover-etal-2006-study} serves as a meaningful proxy for human post-editing effort, complementing and contrasting more advanced neural ``black box'' metrics like COMET \citep{rei-etal-2020-comet}, which was trained via regression to predict human quality judgment scores. We operationally define the predictive power of any given quality metric or heuristic as its Kendall's $\tau$ rank correlation with these two reference scores.

Our analysis yields three primary findings: 
\begin{itemize}
    \item On the source side, the predictive power of difficulty metrics is highly contingent on the ground-truth metric used; features that strongly correlate with COMET (e.g., segment length, neural predictors) show almost no correlation with TER.
    \item On the candidate side, modern QE metrics are significantly more aligned with, and predictive of, the quality of traditional neural MT outputs than those from general-purpose LLMs.
    \item While we confirm a statistically significant positional bias in document-level LLMs (i.e., the tendency for translation quality to degrade for segments occurring later in a document; \citealt{peng:hal-04623006, peng-etal-2025-investigating}) its practical impact on translation quality appears to be negligible.
\end{itemize}
These findings highlight that the architectural shift towards LLMs alters the reliability of established quality prediction methods while simultaneously mitigating previous challenges in document-level translation. To facilitate further research, we make our dataset and code publicly available under CC-BY SA 4.0 license.\footnote{\href{https://github.com/mmarmonier/Hindex}{https://github.com/mmarmonier/Hindex}}

\section{Data}
The experiments in this paper are conducted on a unique dataset that is the byproduct of the creation of the French partition of the OLDI Seed Corpus \citeplanguageresource{marmonier-sagot-bawden:2025:WMT}. We first describe the source corpus and the process used to create the French reference translations, before detailing the resulting multi-candidate dataset used for our analyses.

\subsection{The OLDI Seed Corpus and its French Partition}
The OLDI Seed Corpus (originally the NLLB Seed Corpus) was created to provide a small but high-quality parallel dataset to bootstrap MT capabilities for low-resource languages \citep{nllbteam2022languageleftbehindscaling, maillard-etal-2023-small}. The source material consists of approximately 6,000 English sentences sampled from a curated list of 203 core Wikipedia articles, ensuring broad topic coverage and allowing these segments to be reassembled into documents.

\begin{figure}
    \centering
    \includegraphics[width=0.90\linewidth]{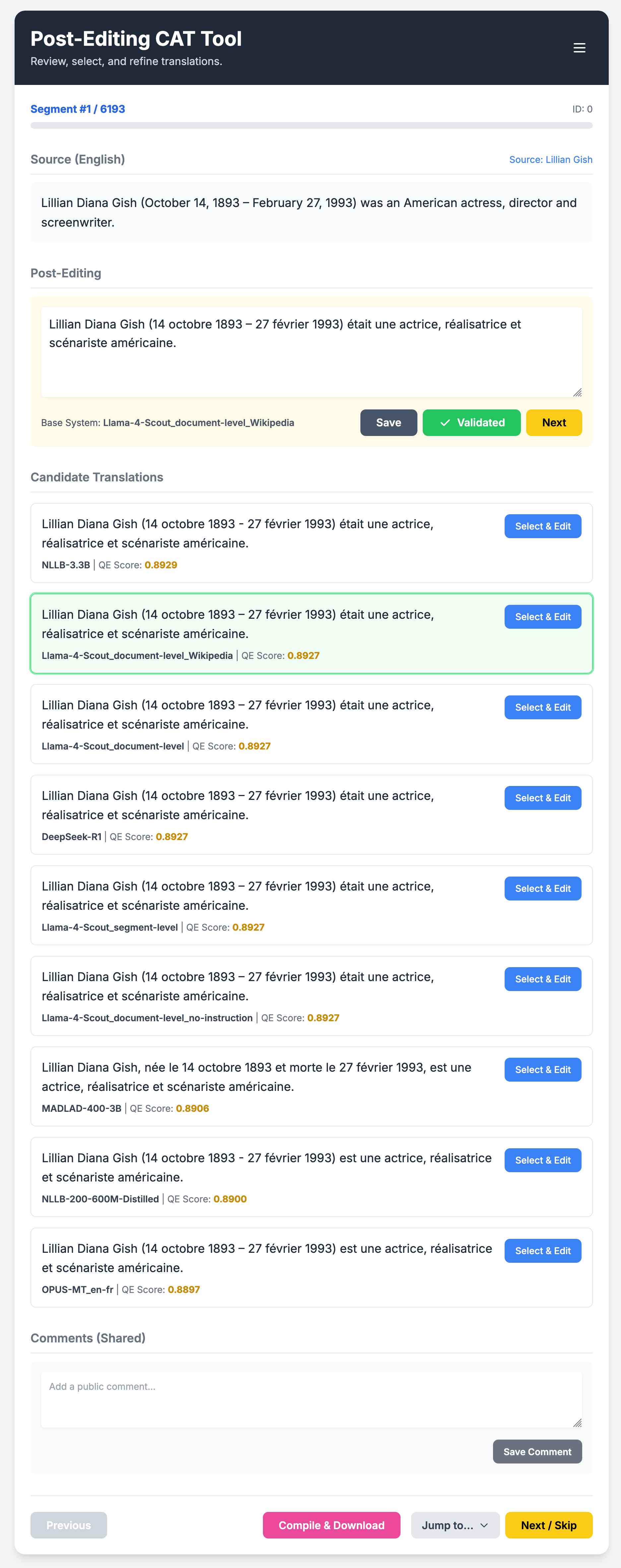}
    \caption{Post-editing interface used to compile the French partition of the OLDI Seed Corpus. Image borrowed from \protect\citeplanguageresource{marmonier-sagot-bawden:2025:WMT}.}
    \label{fig:UI_PE}
\end{figure}

As a contribution to the WMT 2025 Open Language Data Initiative (OLDI) shared task \citep{dale-EtAl:2025:WMT}, a French partition of this corpus was created \citeplanguageresource{marmonier-sagot-bawden:2025:WMT}. The primary goal was to establish a high-quality pivot resource to facilitate the future creation of parallel corpora for the under-resourced regional languages of France whose speakers are far more likely to be fluent in French than in English. Given the high quality of modern English-French MT systems, the French partition was produced using a post-edited MT (MTPE) workflow. This process was carried out by two native French speakers with C2-level proficiency in English, using a custom-built interface (see Figure~\ref{fig:UI_PE}).

\subsection{The Multi-Candidate Byproduct Dataset}
Our current work makes use of the full set of data generated during the creation of the aforementioned French partition. This ``byproduct'' dataset is uniquely suited for our experiments, as for each of the 6,193 English source segments, it contains:

\begin{itemize}
    \item A set of nine different MT hypotheses generated by a diverse array of systems and prompting techniques.
    \item A single, final human post-edited reference translation, which represents the gold-standard output after meticulous review and correction.
\end{itemize}

The nine translation hypotheses were generated using a mix of traditional NMT models and prompted LLMs, which can be grouped into two main categories:

\begin{itemize}
    \item NMT models (\texttt{sent2sent}): Four dedicated sequence-to-sequence Transformer models that translate sentence by sentence. \textbf{OPUS-MT}: A standard bilingual Transformer model trained on open parallel corpora \citep{tiedemann-thottingal-2020-opus}. \textbf{NLLB-3.3B}: The 3.3B parameter multilingual model from the NLLB family \citep{nllbteam2022languageleftbehindscaling}. \textbf{NLLB-600M-Distilled}: The smaller, 600M parameter distilled version of NLLB. \textbf{MADLAD-400-3B}: The 3B parameter multilingual model from the MADLAD-400 project \citeplanguageresource{kudugunta2023madlad400multilingualdocumentlevellarge}.

\item LLMs (\texttt{smallDoc2sent} or \texttt{doc2doc}): Five hypotheses were generated using two LLMs with various prompting strategies. \textbf{Llama-4-Scout\_segment-level}: A 109B-parameter Llama 4 Scout model \citep{meta_ai_llama_2025} translated segments individually, where the prompt included up to five preceding source sentences for local context (\texttt{smallDoc2sent}). \textbf{Llama-4-Scout\_document-level}: The same 109B model translated entire documents in a single pass using a full set of translation guidelines. \textbf{Llama-4-Scout\_document-level\_no-instruction}: A contrastive ablation of the above, where the translation guidelines were removed from the prompt. \textbf{Llama-4-Scout\_document-level\_Wikipedia}: The document-level setup, but with the prompt augmented by the full text of the corresponding French Wikipedia article. \textbf{DeepSeek-R1}: One document-level hypothesis was generated using the 671B-parameter model DeepSeek-R1 \citep{deepseekai2025deepseekr1incentivizingreasoningcapability}.

\end{itemize}

To establish a baseline for the quality of the raw machine-generated hypotheses, we calculated reference-based COMET\footnote{wmt22-comet-da} and TER\footnote{nrefs:1|case:lc|tok:tercom|norm:no|punct:yes| asian:no|version:2.5.1} scores for each of the nine systems against the final human post-edited text.\footnote{We note that while evaluating TER against a single post-edited reference inherently favors the candidate chosen as the starting point, the interface displayed all nine options simultaneously and post-editors actively borrowed from non-selected hypotheses \protect\citeplanguageresource{marmonier-sagot-bawden:2025:WMT}, making TER meaningful across the entire pool.} During data collection, the DeepSeek-R1 model refused to translate a small number of documents (165 segments) pertaining to sensitive topics, due to its safety filters. To ensure a fair comparison, these segments were excluded from all evaluations, leaving a total of 6,028 segments.

\begin{table}[t]
\centering
\tiny
\begin{tabular}{lrrr}
\toprule
System & Mean & 95\% CI & Group \\
\midrule
DeepSeek-R1 & 12.44 & [12.03, 12.86] & A \\
\midrule
Llama-4-Scout\_segment-level & 26.09 & [25.63, 26.55] & B \\
MADLAD-400-3B & 26.66 & [26.19, 27.13] & B \\
Llama-4-Scout\_document-level\_no-instruction & 26.68 & [26.22, 27.15] & B \\
Llama-4-Scout\_document-level & 26.73 & [26.26, 27.19] & B \\
Llama-4-Scout\_document-level\_Wikipedia & 27.27 & [26.75, 27.78] & B \\
NLLB-3.3B & 27.73 & [27.26, 28.21] & B \\
\midrule
OPUS-MT\_en-fr & 31.80 & [31.28, 32.32] & C \\
NLLB-200-600M-Distilled & 32.75 & [30.92, 34.57] & C \\
\bottomrule
\end{tabular}
\caption{\textbf{TER} reference scores with 95\% confidence intervals (\(\downarrow\) lower is better).}
\label{tab:ter_reference_scores}
\end{table}

\begin{table}[t] \centering \tiny \begin{tabular}{lrrr} \toprule System & Mean & 95\% CI & Group \\ \midrule DeepSeek-R1 & 0.9219 & [0.9206, 0.9232]  & A \\ \midrule Llama-4-Scout\_segment-level & 0.8982 & [0.8968, 0.8997]  & B \\ MADLAD-400-3B & 0.8956 & [0.8940, 0.8971]  & B \\ Llama-4-Scout\_document-level\_no-instruction & 0.8954 & [0.8938, 0.8969]  & B \\ Llama-4-Scout\_document-level & 0.8949 & [0.8933, 0.8964]  & B \\ NLLB-3.3B & 0.8934 & [0.8918, 0.8949]  & B \\ Llama-4-Scout\_document-level\_Wikipedia & 0.8917 & [0.8898, 0.8936]  & B \\ \midrule NLLB-200-600M-Distilled & 0.8779 & [0.8761, 0.8798]  & C \\ \midrule OPUS-MT\_en-fr & 0.8730 & [0.8709, 0.8750]  & D \\ \bottomrule \end{tabular} \caption{\textbf{COMET} reference scores with 95\% confidence intervals (\(\uparrow\) higher is better).} \label{tab:comet_reference_scores} \end{table}

Tables~\ref{tab:ter_reference_scores} and \ref{tab:comet_reference_scores} present the average performance of each system. The results show a clear hierarchy of quality. DeepSeek-R1 stands out as the top-performing system, achieving significantly higher COMET and lower TER scores than all other models. The remaining systems fall into several distinct statistical groups, with the larger NMT models and all Llama 4 prompting variants forming a competitive middle tier, and the smaller OPUS-MT and distilled NLLB models occupying the lower performance tiers.

\section{Source-Side Experiments}
Our first set of experiments investigates source-side quality prediction, more commonly known as ``translation difficulty estimation.'' The aim is to predict the eventual quality of a translation or, conversely, the post-editing effort it will require, a priori, without access to the translation itself, relying exclusively on the source segment.

\subsection{Methodology}
To assess the predictive power of various source-side features, we measure their correlation with the final, reference-based translation quality measured by COMET or TER. We evaluate twelve metrics drawn from different research paradigms, which can be categorized as follows:

    \begin{itemize}
        \item \textbf{Linguistic and Readability Metrics}: Following foundational approaches to this task \citep{mishra-etal-2013-automatically}, we include nine traditional metrics; five standard readability formulas (Dale-Chall, SMOG index, Flesch Reading Ease, Coleman-Liau index, and Flesch-Kincaid Grade) and four linguistic complexity features (segment length, word rarity, degree of polysemy, and syntactic tree height). See Section \ref{sec:appendix_readability} in the appendix for details.
        \item \textbf{Neural Metrics}: State-of-the-art performance is now typically achieved with dedicated neural models \citep{fernicola-etal-2023-return, proietti2025estimating}. We include two such specialized predictors from the Sentinel family (sentinel-src-da and sentinel-src-mqm) \citep{proietti2025estimating}. Another prominent neural approach, which conceptually borders on the candidate-side, gauges difficulty via ``glass box'' capture of a translation model's surprisal \citep{lim-etal-2024-predicting}; to operationalize this as a single source-side heuristic for benchmarking, we calculated the MT surprisal for the NLLB-200 600M model on its own translations only, and assigned these scores to the source segments.
    \end{itemize}

For our analysis, we use Kendall's rank correlation coefficient ($\tau_b$) to measure the association between each source-side metric and the ground-truth quality scores (TER and COMET). Kendall's $\tau_b$ is a robust non-parametric test that measures the strength of association between two variables based on the similarity of their rankings. It functions by comparing the number of concordant pairs (pairs of segments that are in the same relative order for both the difficulty metric and the quality score) and discordant pairs (pairs that are in the opposite order). We conducted the analysis per system. Specifically, for each of the 6,028 segments, we ranked the list of source-side metric scores (e.g., \textit{segment\_length}) and compared this ranking against the corresponding list of ground-truth quality scores (TER or COMET) for a single translation system (e.g., NLLB-3.3B). This process was then repeated for every source metric and every system. Unlike Pearson's r, which assumes a linear relationship and normally distributed data, rank-based coefficients are more appropriate for MT quality scores, whose distributions often violate these assumptions. We prefer Kendall's $\tau_b$ over Spearman's $\rho$ as it is generally considered more robust and reliable in the presence of tied ranks, not uncommon in this type of data.

\subsection{Results and Analysis}
\begin{figure}
    \centering
    \includegraphics[width=1\linewidth]{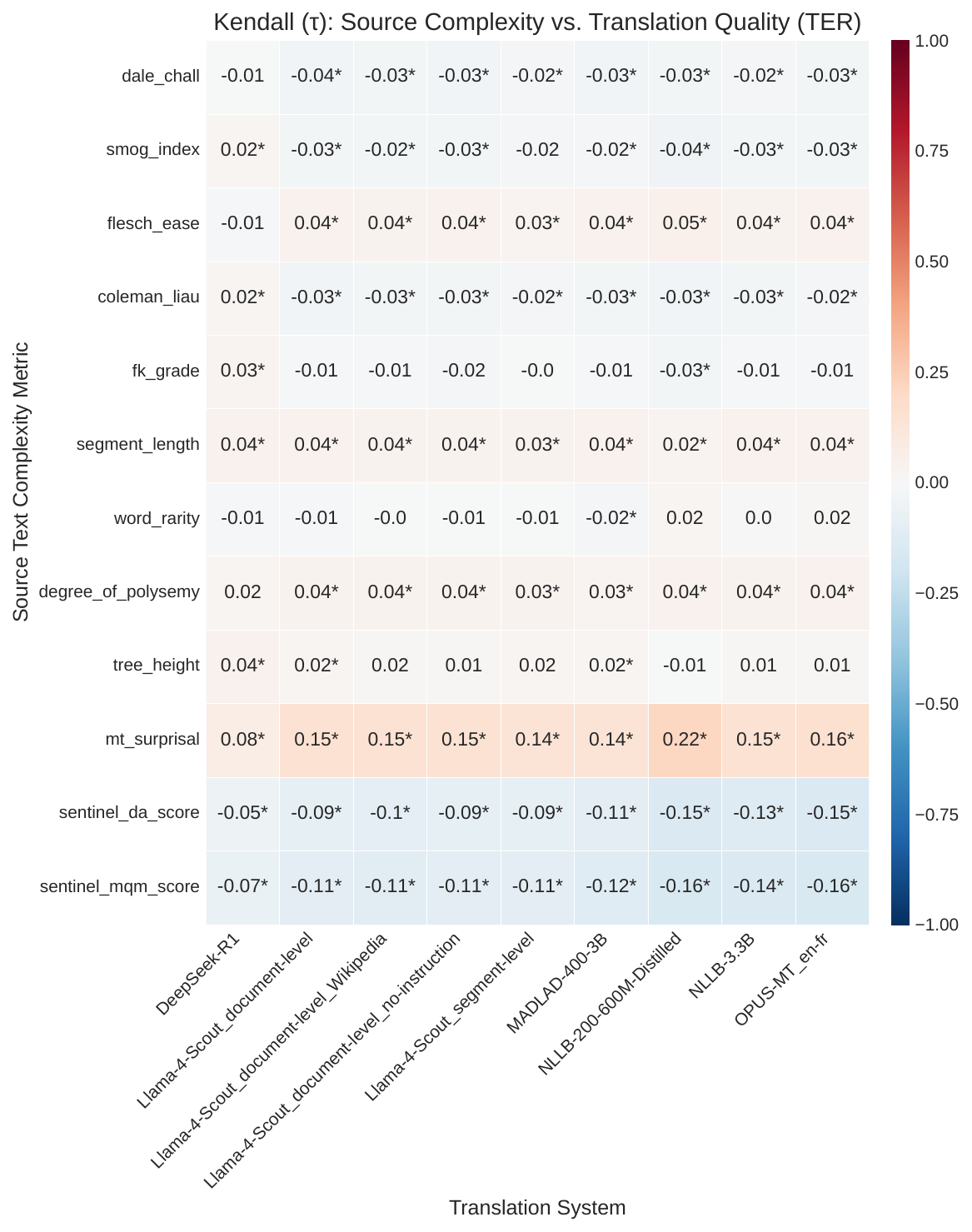}
    \caption{Kendall's~$\tau$ correlation between source-side metrics and translation quality as measured by TER. 
An asterisk~($^*$) indicates a statistically significant correlation ($p < 0.05$).}
    \label{fig:ter_aD}
\end{figure}

\begin{figure}
    \centering
    \includegraphics[width=1\linewidth]{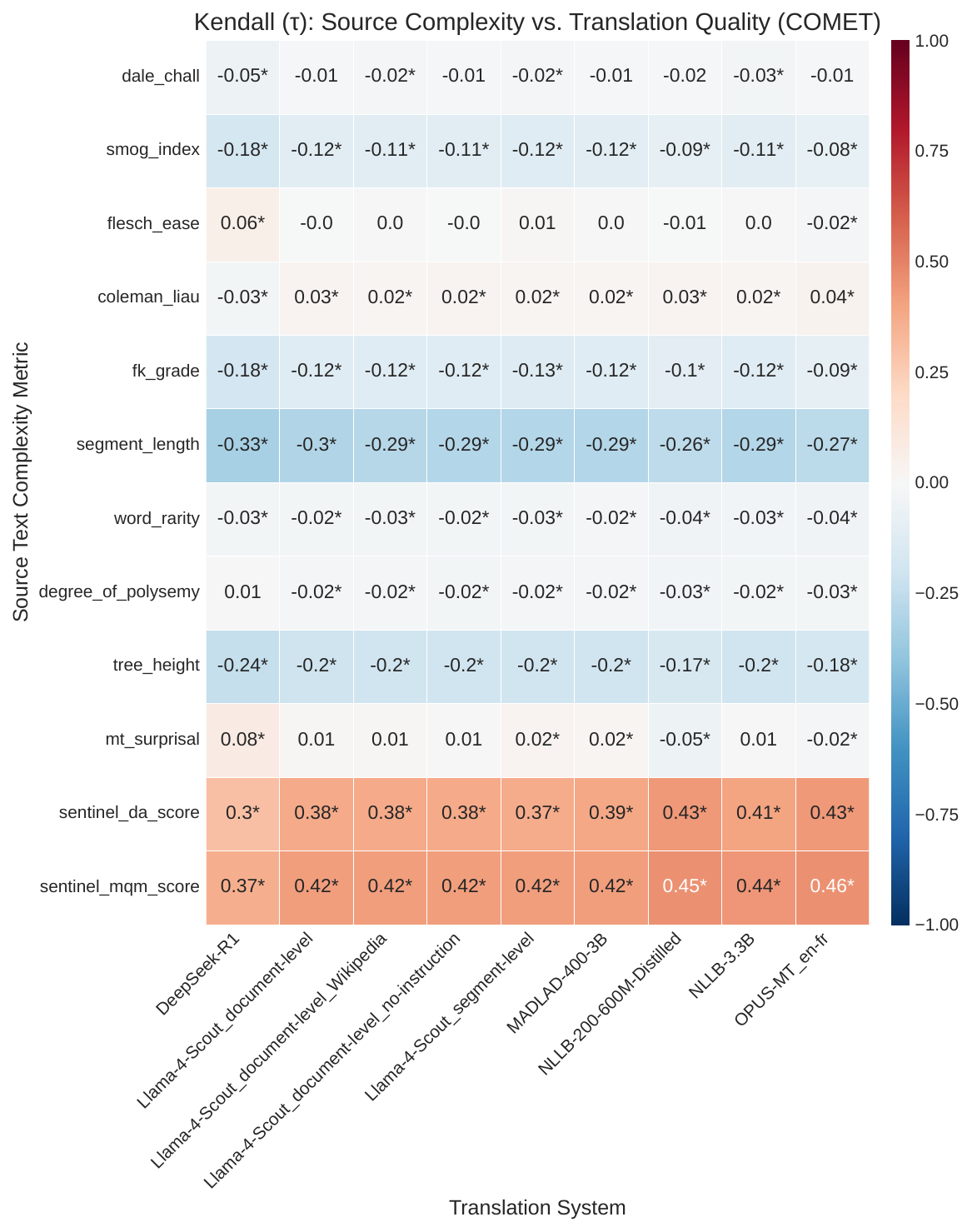}
    \caption{Kendall's~$\tau$ correlation between source-side metrics and translation quality as measured by COMET. 
An asterisk~($^*$) indicates a statistically significant correlation ($p < 0.05$). 
}
    \label{fig:comet_aD}
\end{figure}

The results of our correlation analysis are presented in Figures~\ref{fig:ter_aD} and \ref{fig:comet_aD}. A key finding is that the predictive power of source-side metrics is dependent on whether translation quality is measured with TER (post-editing effort) or COMET (regression on Direct Assessment scores from past WMT shared tasks).

When quality is measured by TER (Figure~\ref{fig:ter_aD}) the correlation patterns are generally modest. The most consistent positive predictor is \textit{mt\_surprisal} ($\tau$ between $0.14$ and $0.16$ across most systems). Notably, this surprisal---calculated here using the NLLB-200-distilled model for which it should therefore not be interpreted as a strictly source-side metric---correlates with the TER scores of translations from other systems, including the much larger MADLAD-400-3B and Llama-4. Other metrics show very weak correlations to TER. The Sentinel models, which are strong predictors of COMET, show a much weaker negative correlation with this edit rate ($\tau$ between $-0.05$ and $-0.16$), and \textit{segment\_length} and \textit{tree\_height} show almost no correlation with post-editing effort.

In contrast, when quality is measured by COMET (Figure~\ref{fig:comet_aD}), the predictive patterns are markedly different and the correlations appear much stronger. Here, the Sentinel neural metrics (\textit{sentinel\_da\_score} and \textit{sentinel\_mqm\_score}) are by far the most effective predictors, exhibiting consistent positive correlations across all systems ($\tau$ from $0.30$ to $0.46$). Similarly, \textit{segment\_length} ($\tau$ between $-0.26$ and $-0.33$) and \textit{tree\_height} ($\tau$ between $-0.17$ and $-0.24$) now show the most salient negative correlations. 

We hypothesize that this apparent strength in the COMET-based correlations is attributable in part to system artifacts. (i)~Data and Architectural Overlap: Both the Sentinel predictors and the COMET reference metric are derived from the same XLM-R-based architectures and are likely trained on overlapping data from WMT shared tasks. Their high correlation may therefore reflect this shared foundation rather than independent predictive power. (ii)~Metric-Internal Bias: While \textit{segment\_length} and \textit{tree\_height} show weak correlation with TER, they are strong positive predictors for COMET. COMET might have derived a misleading heuristic from its training data: since longer source segments give more opportunity for errors in translation candidates that subsequently receive lower DA scores from human annotators, the metric might have learned to associate source length directly with low translation quality. TER does not reflect this bias, suggesting that length itself was not a primary driver of proportional post-editing effort.

Whatever its cause, this dichotomy demonstrates that the source-side features that best predict translation difficulty are contingent upon the specific reference-based metric used to define said quality.

\section{Candidate-Side Experiments}
Our second set of experiments shifts the focus from the source text to the translation candidates to address two key aspects of quality prediction: the predictive power of standard Quality Estimation (QE) metrics and of positional bias in LLM-based document-level MT.

We first conduct a ``hindsight'' QE evaluation. We assess the ability of two state-of-the-art reference-free QE metrics, COMET-QE\footnote{wmt22-cometkiwi-da} \citep{rei-etal-2022-cometkiwi} and MetricX-QE\footnote{metricx-24-hybrid-xl-v2p6} \citep{juraska-etal-2024-metricx}, to predict the final quality of the nine system outputs. Their predictions are correlated against our gold-standard reference scores (TER and COMET) to measure their performance across systems. Second, we investigate positional bias, a known challenge \citep{peng:hal-04623006, peng-etal-2025-investigating} where translation quality degrades as a model processes a long document.

\subsection{Hindsight Quality Estimation}
 The original data collection interface (cf. Figure~\ref{fig:UI_PE}) was not anonymized; it displayed both the system name and a COMET-Kiwi (QE) score for each of the nine candidates. This introduces a strong potential for anchoring bias, as post-editors might have been influenced by the visible QE score when selecting a hypothesis to edit. Surprisingly, however, the data strongly suggests that post-editors were not captive to this bias. Table~\ref{tab:comet_qe_analysis} shows the average rank of each system as determined by the COMET-Kiwi QE score from the original dataset. This ranking does not align with the final, reference-based quality rankings from Tables~\ref{tab:ter_reference_scores} and \ref{tab:comet_reference_scores}. For instance, DeepSeek-R1, the clear top system by both TER and COMET, was only ranked 5th on average by the QE model. Conversely, NLLB-3.3B was ranked 2nd by the QE model but was a mid-to-low performer in the final quality assessment. This discrepancy indicates that the human post-editors frequently overrode the QE metric's ranking to select a candidate they deemed a better starting point.

\begin{table*}[h!]
\centering
\scriptsize
\begin{tabular}{lcccccccc}
\toprule
\textbf{System Name} & \textbf{Avg Rank} & \textbf{\% 1st} & \textbf{\# 1st} & \textbf{\% 1st Solo} & \textbf{\# 1st Solo} & \textbf{\% Last} & \textbf{\# Last} \\
\midrule
Llama-4-Scout\_document-level\_no-instruction & 4.12 & 14.58\% & 903 & 1.63\% & 101 & 12.18\% & 754 \\
NLLB-3.3B & 4.12 & 19.62\% & 1215 & 10.45\% & 647 & 10.04\% & 622 \\
MADLAD-400-3B & 4.13 & 21.36\% & 1323 & 13.00\% & 805 & 11.06\% & 685 \\
Llama-4-Scout\_document-level & 4.18 & 14.53\% & 900 & 1.37\% & 85 & 12.98\% & 804 \\
DeepSeek-R1 & 4.27 & 23.53\% & 1457 & 16.83\% & 1042 & 15.52\% & 961 \\
Llama-4-Scout\_segment-level & 4.35 & 18.57\% & 1150 & 11.30\% & 700 & 13.69\% & 848 \\
Llama-4-Scout\_document-level\_Wikipedia & 4.44 & 14.06\% & 871 & 4.25\% & 263 & 14.79\% & 916 \\
NLLB-200-600M-Distilled & 5.24 & 14.58\% & 903 & 8.96\% & 555 & 21.73\% & 1346 \\
OPUS-MT\_en-fr & 5.47 & 15.36\% & 951 & 10.17\% & 630 & 27.92\% & 1729 \\
\bottomrule
\end{tabular}
\caption{Analysis of COMET-Kiwi (QE) score distribution displayed to post-editors. Shows the average rank (lower is better), percentage/count of segments where a system was ranked 1st, percentage/count where it was ranked 1st with no ties (\% 1st Solo), and percentage/count where it was ranked last.}
\label{tab:comet_qe_analysis}
\end{table*}

\begin{figure*}[t]
\centering
\begin{subfigure}[b]{\linewidth}
  \includegraphics[width=\linewidth]{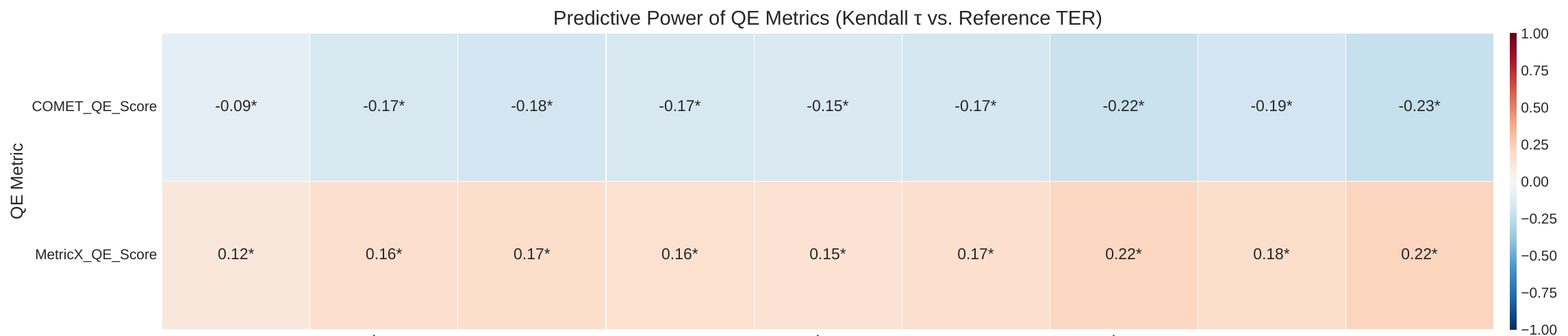}
  \label{fig:sub1b}
\end{subfigure}


\begin{subfigure}[b]{\linewidth}
  \includegraphics[width=\linewidth]{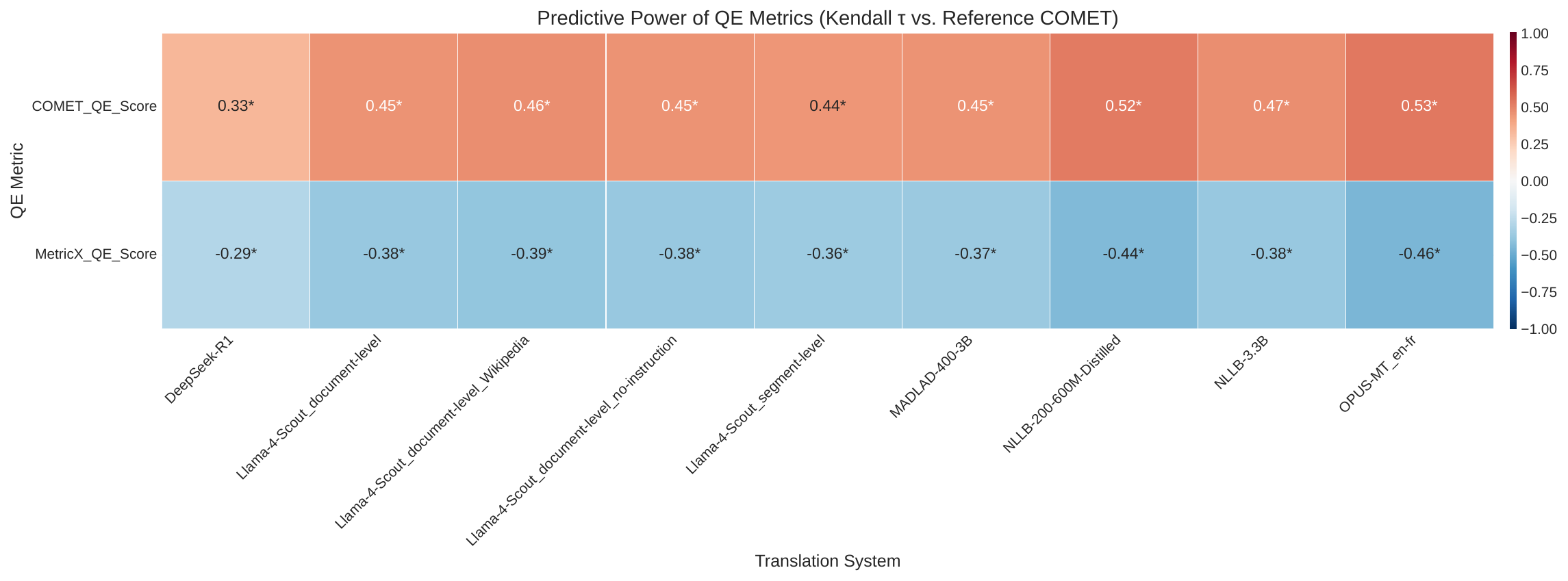}
  \label{fig:sub2b}
\end{subfigure}

\caption{Kendall's $\tau$ correlation between two reference-free QE metrics (\textit{COMET\_QE}, \textit{MetricX\_QE}) and our gold-standard reference scores. The top plot shows correlations against reference TER, while the bottom plot uses reference COMET. An asterisk (*) indicates statistical significance ($p<0.05$).}
\label{fig:QEperSystem}
\end{figure*}

Our hindsight QE evaluation confirms this hypothesis. The results are summarized in Figure~\ref{fig:QEperSystem}. The predictive power of the QE metrics is found to not be uniform across systems, as both QE metrics correlate more strongly with the quality of the weaker MT systems than with the top-performing one. For instance, when predicting reference COMET, the correlation for \textit{COMET\_QE} is considerably higher for one of the weakest systems, OPUS-MT ($\tau=0.53$), than for the strongest, DeepSeek-R1 ($\tau=0.33$). This suggests that current QE models are better at identifying errors in mediocre translations than at distinguishing fine-grained quality differences among high-quality candidates. We hypothesize, based on observations made by the post-editors of the French OLDI Seed partition, that this weakness is rooted in a lack of the factual knowledge required to judge terminological accuracy in the encyclopedic domain to which the OLDI Seed Corpus belongs.

We note again that the high correlation between \textit{COMET\_QE} and reference COMET is likely to be an expected artifact of their shared base model (derived from XLM-R) and overlapping training data.\footnote{The same observation applies to MetricX, though to a lesser extent.} A more telling benchmark is their performance against reference TER, where the same behavior is observed.

\begin{figure}[t] 
\centering

\begin{subfigure}[b]{0.48\columnwidth}
  \includegraphics[width=\linewidth]{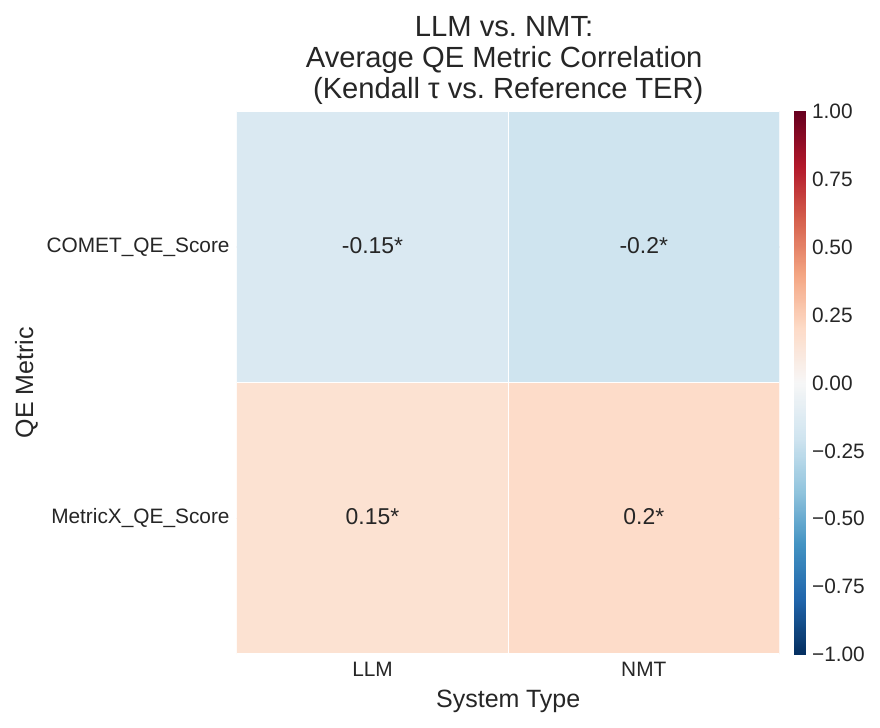}
  \caption{}
  \label{fig:sub1c}
\end{subfigure}
\hfill
\begin{subfigure}[b]{0.48\columnwidth}
  \includegraphics[width=\linewidth]{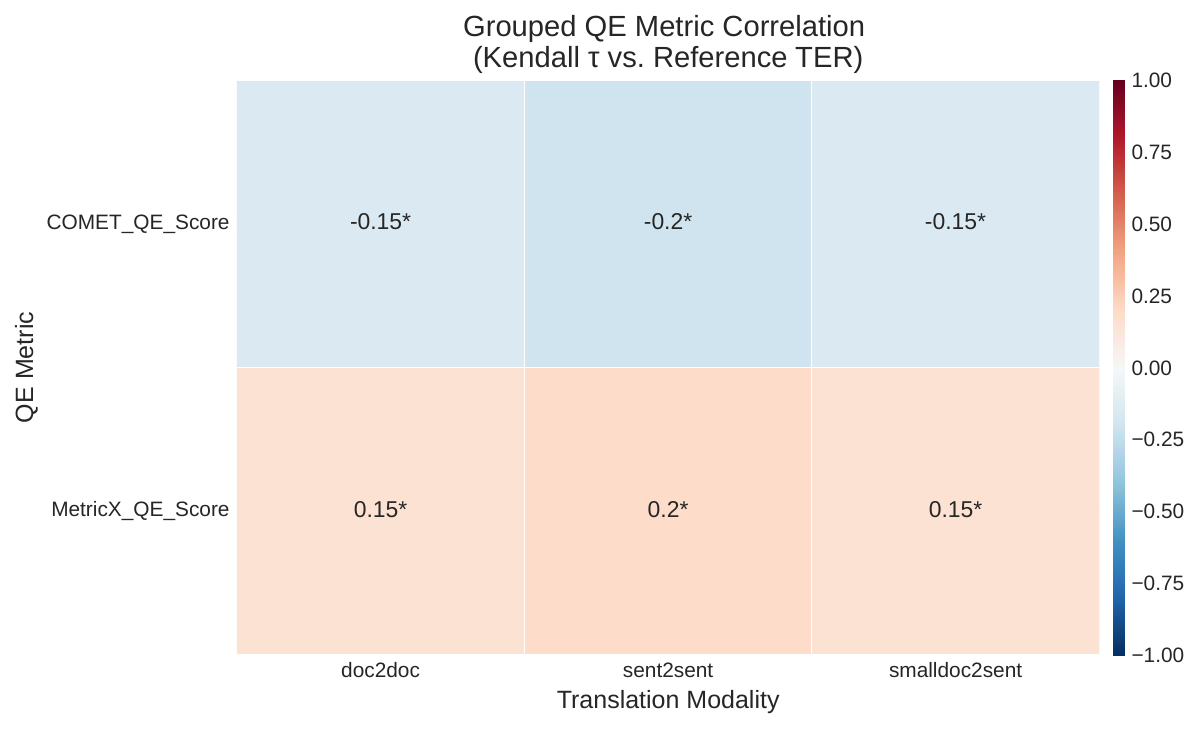}
  \caption{}
  \label{fig:sub2c}
\end{subfigure}

\vspace{0.5em} 

\begin{subfigure}[b]{0.48\columnwidth}
  \includegraphics[width=\linewidth]{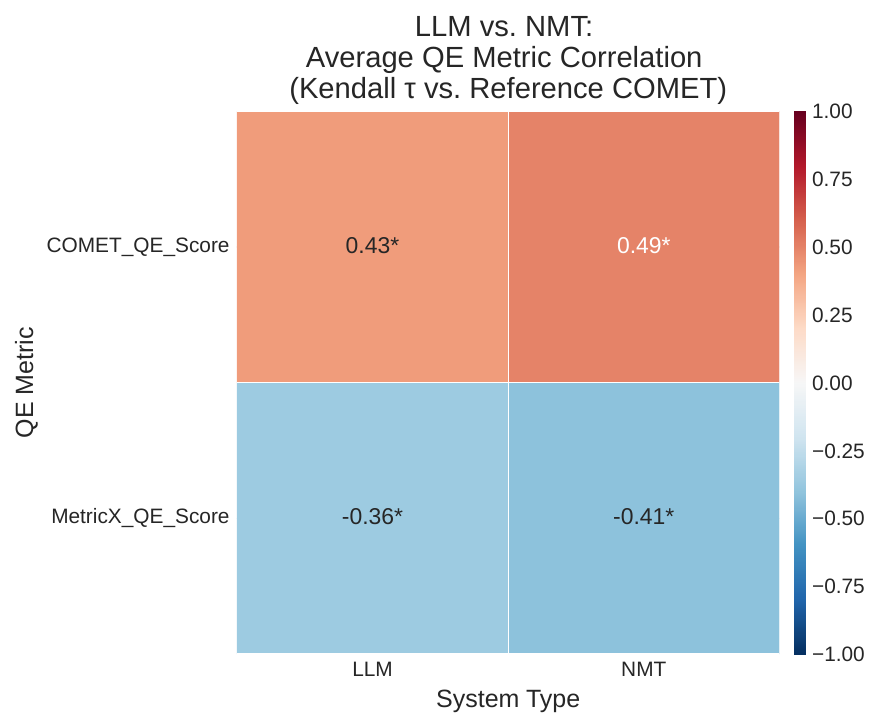}
  \caption{}
  \label{fig:sub3}
\end{subfigure}
\hfill
\begin{subfigure}[b]{0.48\columnwidth}
  \includegraphics[width=\linewidth]{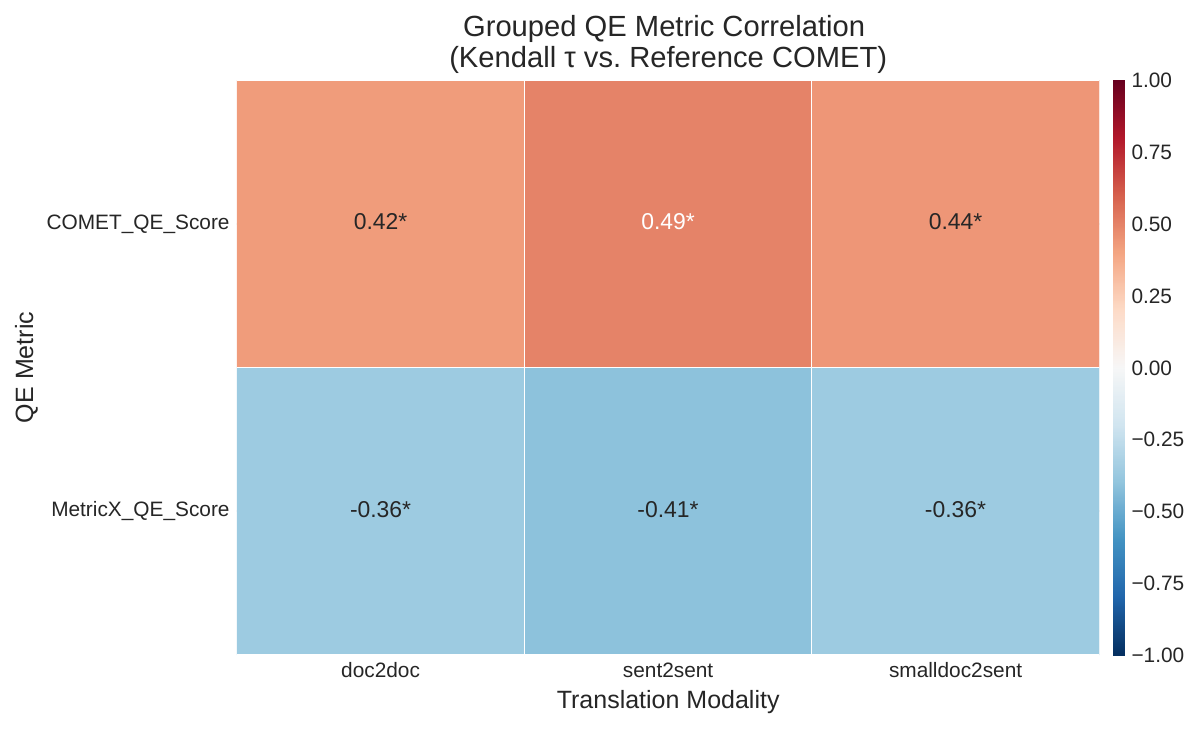}
  \caption{}
  \label{fig:sub4}
\end{subfigure}

\caption{Kendall's $\tau$ correlation between QE metrics and gold-standard scores, averaged across system groups. The top row shows correlations against reference TER; the bottom row uses reference COMET. The left column groups by system type (LLM vs. NMT), while the right column groups by translation granularity.}
\label{fig:quadrant}
\end{figure}

When the results are aggregated by system type as in Figure~\ref{fig:quadrant}, an explanatory pattern emerges: the predictive power of the QE metrics is consistently stronger for the traditional NMT models than for the LLM-based systems. For example, \textit{COMET\_QE}'s average correlation with reference COMET is highest for the NMT group ($\tau=0.49$), compared to the LLM group ($\tau=0.43$). This significant trend\footnote{We confirmed the statistical significance of these differences in correlation strength using bootstrap resampling ($p<0.05$).} holds across all conditions, as \textit{MetricX\_QE\_Score} also shows its strongest correlation with reference TER for the NMT group ($\tau=0.20$ vs. $\tau=0.15$ for the LLM group).

The analysis by translation granularity suggests that this is an architectural effect, not simply a matter of context size. The \texttt{smallDoc2sent} LLM, despite its limited taxing of the context window, patterns with the \texttt{doc2doc} LLMs rather than the \texttt{sent2sent} NMTs. This finding suggests that the predictions of current QE models are more closely aligned with the outputs of specialized NMT architectures than with those of general-purpose LLMs, explaining why post-editors frequently overrode QE recommendations to select LLM hypotheses as starting points.

\subsection{Positional Bias in Document-Level Translation}

Our final experiment investigates positional bias in document-level MT, a phenomenon where a segment's translation quality degrades the later it appears within a document, and its candidate-side predictive power.

To measure a segment's position from a model-centric perspective, we calculated its cumulative token rank: the total number of tokens (from the prompt, thought tokens in the case of DeepSeek-R1, and prior generation) processed by a given \texttt{doc2doc} model before it began translating the segment in question, with each model's specific tokenizer. We then correlated this rank with translation quality.

\begin{figure}[t]
\centering
\begin{subfigure}[b]{\columnwidth}
  \includegraphics[width=\linewidth]{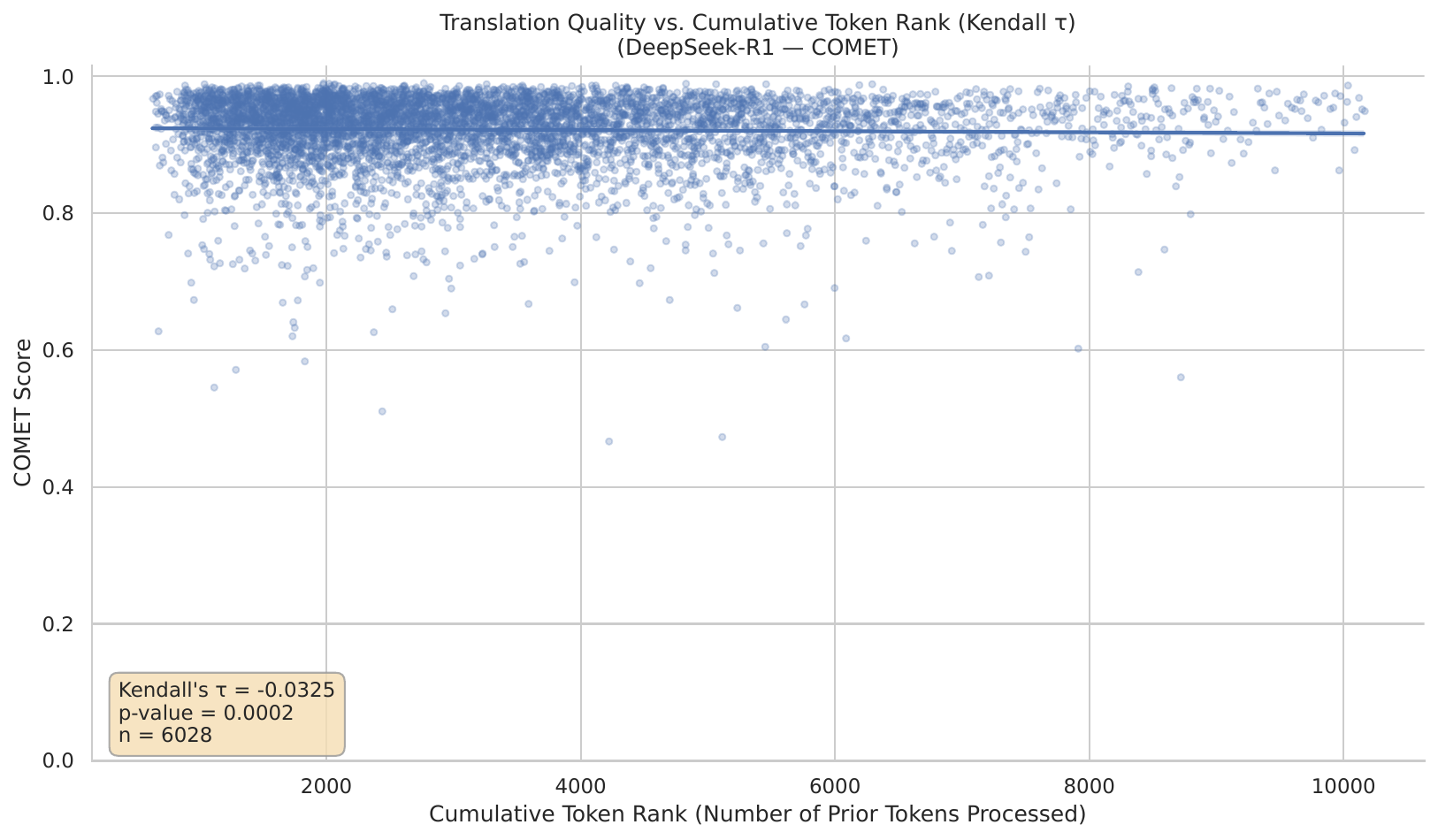}
  \label{fig:sub1d}
\end{subfigure}


\begin{subfigure}[b]{\columnwidth}
  \includegraphics[width=\linewidth]{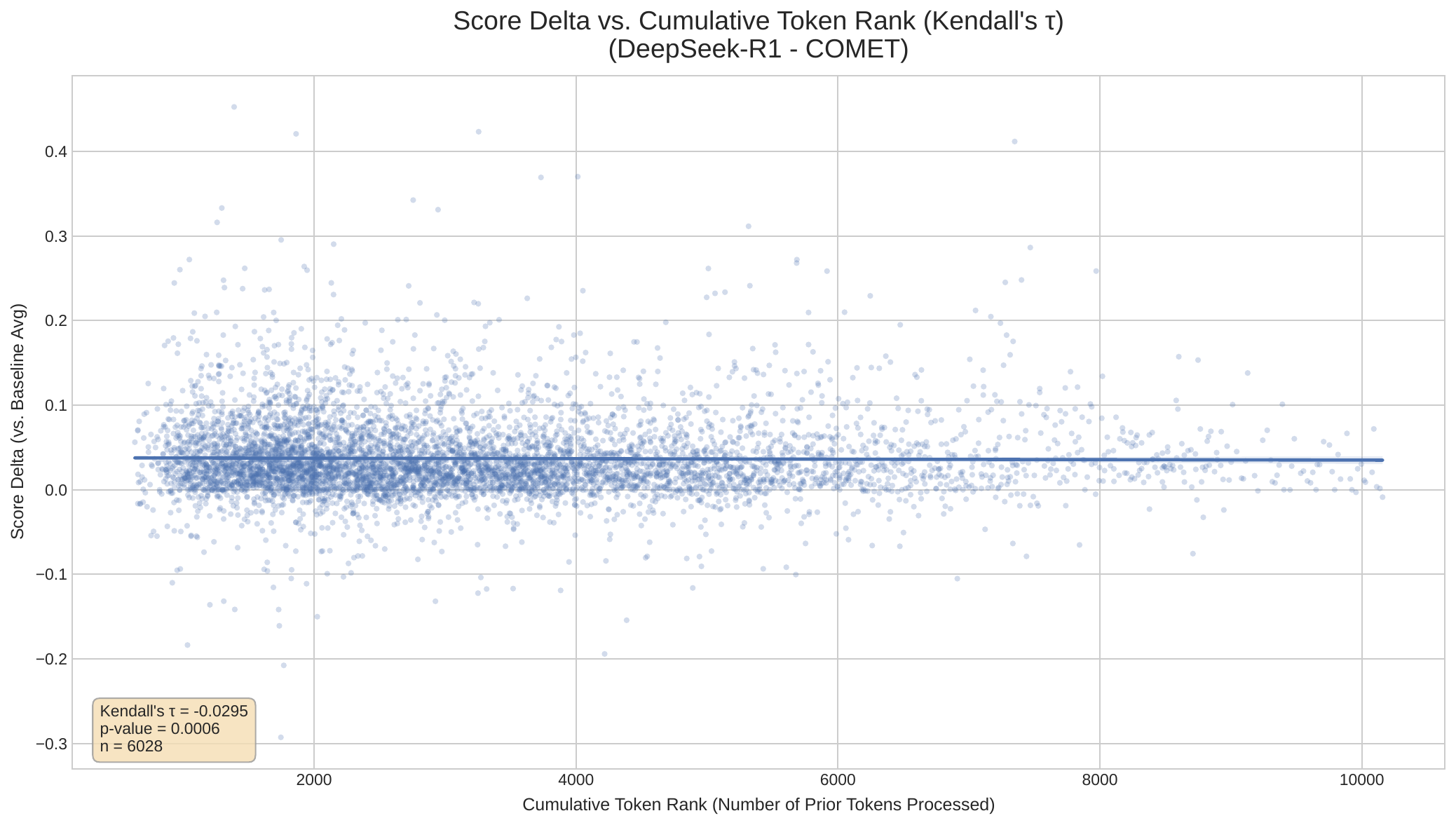}
  \label{fig:sub2d}
\end{subfigure}

\caption{Scatter plots illustrating the effect of positional bias on the DeepSeek-R1 system. The top plot shows the weak negative correlation between raw COMET scores and the cumulative token rank. The bottom plot uses our normalized ``delta score,'' which controls for source-text difficulty.}
\label{fig:scatterplots}
\end{figure}

Correlating this rank directly with raw quality scores reveals a very weak but statistically significant negative trend (e.g., Figure~\ref{fig:scatterplots}, top; $\tau=-0.0325$ for DeepSeek-R1). However, this simple correlation is hard to interpret. The inherent difficulty of the source text acts as a major confounding factor, making it difficult to disentangle the effect of position from the effect of the source material itself.

To isolate the impact of position, we normalize the scores by calculating a ``delta score'': the difference between a \texttt{doc2doc} system's score and the average score of all baseline \texttt{sent2sent} systems for the same segment. This score delta better represents the performance difference attributable specifically to the \texttt{doc2doc} processing strategy, providing a clearer signal of positional degradation. Figure~\ref{fig:scatterplots} illustrates this process for the DeepSeek-R1 system and its COMET scores. The raw COMET scores show a weak but statistically significant negative correlation with token rank ($\tau = -0.0325, p = 0.0002$), confirming the presence of a positional bias. After applying our normalization, this negative trend persists ($\tau = -0.0295, p = 0.0006$), demonstrating that the performance degradation is not merely an artifact of source difficulty but is linked to the segment's position within the document.

\begin{figure*}[t]
\centering
\begin{subfigure}[b]{1\columnwidth}
  \includegraphics[width=\linewidth]{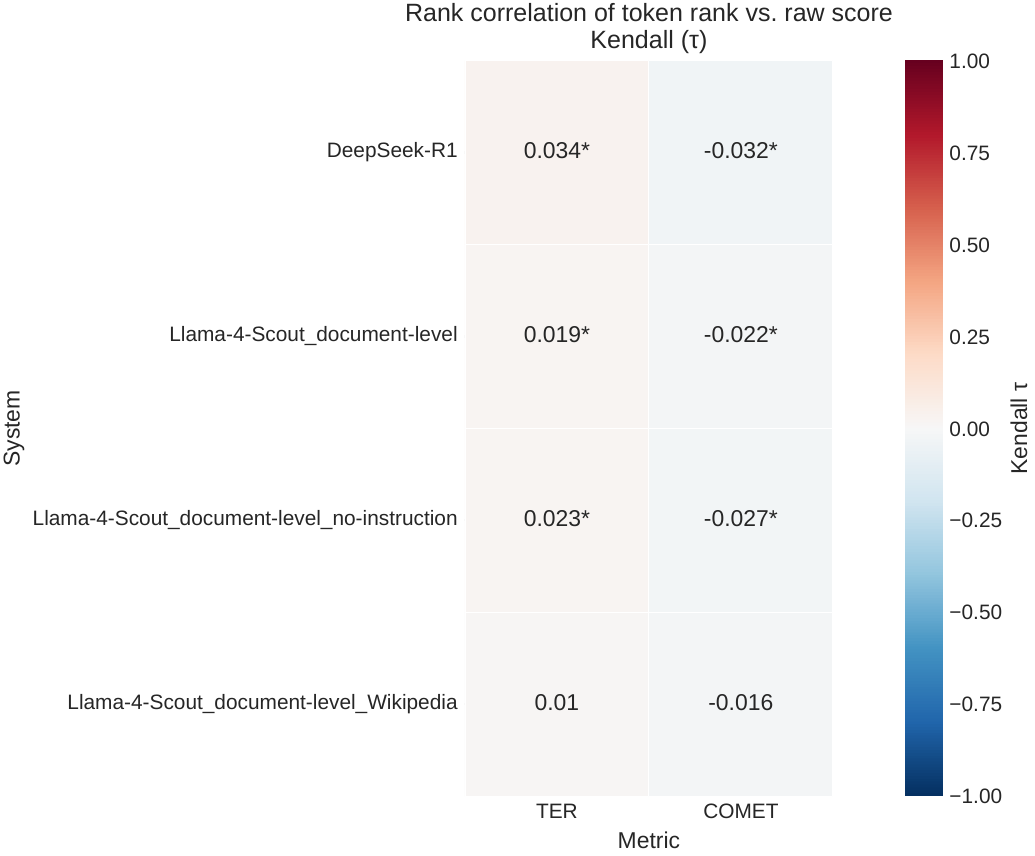}
  \caption{}
  \label{fig:sub1a}
\end{subfigure}
\hfill
\begin{subfigure}[b]{1\columnwidth}
  \includegraphics[width=\linewidth]{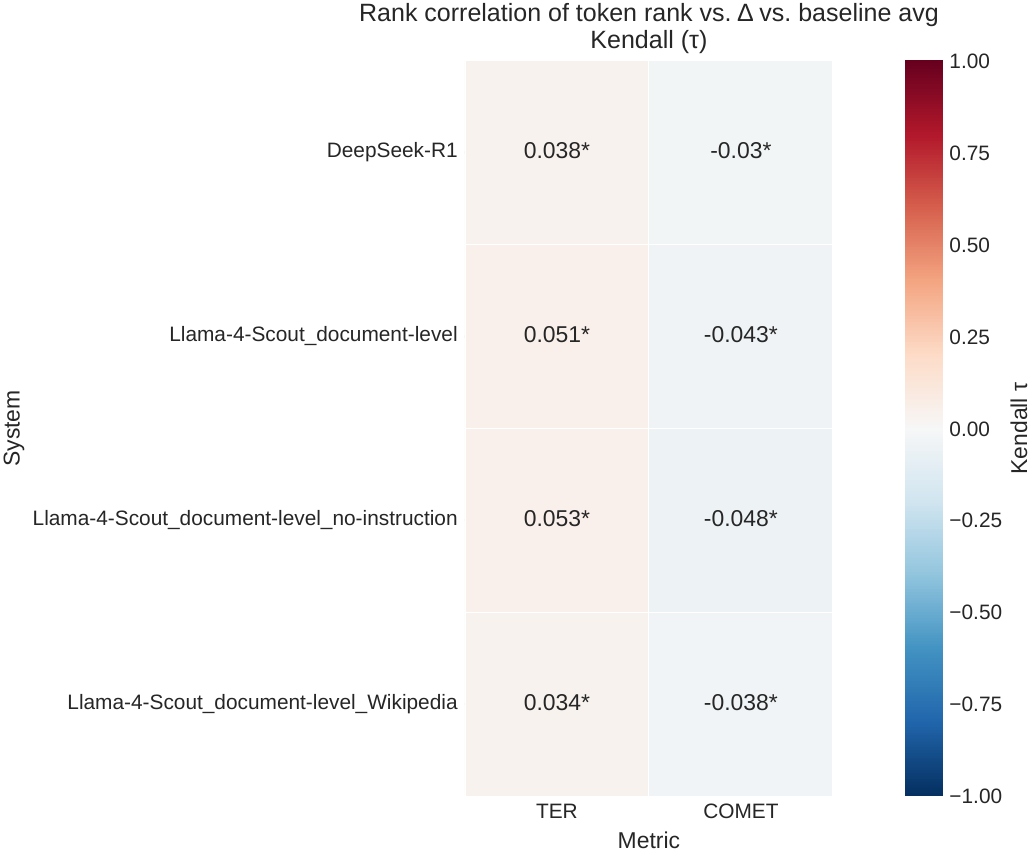}
  \caption{}
  \label{fig:sub2a}
\end{subfigure}
\caption{Kendall's $\tau$ correlation between a segment's cumulative token rank and its quality for all \texttt{doc2doc} systems. Plot (a) uses raw quality scores, while plot (b) uses our normalized delta scores.}
\label{fig:combined_bias}
\end{figure*} 

The full results of this positional bias analysis are presented in Figure~\ref{fig:combined_bias}. The normalization reveals two key insights into the behavior of the different \texttt{doc2doc} systems. First, for the Llama-4-Scout model settings, applying the delta-score normalization noticeably strengthens the negative correlation between position and quality. For example, the COMET score correlation for the base document-level model nearly doubles, from $\tau=-0.022$ to $\tau =-0.043$. This confirms that for these systems, a subtle positional bias is indeed present and is more clearly isolated once the confounding effect of source difficulty is removed. In contrast, the correlation for the top-performing DeepSeek-R1 remains remarkably stable ($\tau \approx -0.03$) with or without normalization. This suggests that DeepSeek-R1's performance is less sensitive to the intrinsic difficulty of the source text to begin with; its robustness to source-side challenges means that the normalization has little effect.

However, the most critical finding across all systems is that the absolute magnitude of the correlation remains very low ($|\tau| < 0.05$). While we can confirm that positional bias is a statistically detectable phenomenon, its practical impact appears to be minimal for this task, suggesting it does not represent a bottleneck for document-level translation quality anymore with current advanced, long-context models.

\section{Related Work}
Early efforts at predicting translation quality from the source text alone framed the task as estimating translation difficulty, relying on readability metrics (used in translation studies research for the same purpose), linguistic features such as sentence length, degree of polysemy, and syntactic complexity to approximate the cognitive effort and expected error rates of human translators \citep{mishra-etal-2013-automatically}. Related work has recently shifted from these feature-based methods toward supervised neural models. This new paradigm involves training encoders to regress directly on human quality judgments, like Direct Assessment (DA; cf. \citealt{graham-etal-2013-continuous}) or MQM scores \citep{burchardt-2013-multidimensional}, using only the source text as input \citep{fernicola-etal-2023-return, proietti2025estimating}. Other neural methods have also proven effective, for instance by showing that MT surprisal strongly correlates with human-assessed translation difficulty \citep{lim-etal-2024-predicting}.

On the candidate side, QE evolved from early work on confidence estimation for MT \citep{ueffing-etal-2003-confidence, gandrabur-foster-2003-confidence, blatz-etal-2004-confidence} into a formal research program with the launch of the WMT QE shared task in 2012 \citep{callison-burch-etal-2012-findings}, with successive WMT editions broadening its scope. The field has now shifted towards systems that directly regress to DA scores, rather than predicting post-editing effort, enabling the pre-training of QE models on annotator data from WMT's shared tasks \citep{rei-etal-2020-comet}.

The shift towards document-level MT) has reintroduced challenges related to input length that affect translation quality \citep{10.1145/3441691}. While general length biases, where performance degrades on documents longer than those seen during training, are a known issue \citep[][\textit{inter alia}]{zhuocheng-etal-2023-addressing}, a more subtle problem is that of position bias. Recent studies have shown that the quality of translation can decrease for sentences appearing later in a document \citep{peng:hal-04623006, peng-etal-2025-investigating}, suggesting that a sentence's absolute position within the input sequence impacts its final translation. This issue echoes challenges from early neural MT at the segment level, where strategies such as reversing the source sentence were developed to address position-dependent performance issues in segment-level models \citep{10.5555/2969033.2969173}. These issues were solved by the advent of attentive models \citep{BahdanauChoBengio2015ICLR}.

\section{Conclusion}
This paper presented a series of quality prediction experiments using a unique dataset generated from a real-world, multi-candidate post-editing workflow. Our ``hindsight'' analyses  yield three primary findings. 

First, on the source side, we find that the predictive power of translation difficulty metrics is contingent on the ground-truth metric used to measure quality. Heuristics like segment length and the neural Sentinel predictors show strong correlations with COMET. However, we hypothesize that these are in part artifacts stemming from shared model architectures and internal metric biases, as these same features show weaker to no correlation with TER, our proxy for post-editing effort. 

Second, on the candidate side, our analysis of the original post-editing process reveals a significant mismatch between the QE model rankings provided to the annotators and their final, human-adjudicated quality, demonstrating that post-editors actively overrode the QE model's guidance. This indicates that reference-free QE metrics are not uniformly aligned with human judgment. We further show that their predictions are more strongly aligned with traditional NMT architectures than with general-purpose LLMs, particularly for top-performing systems, a finding which may help explain this observed discrepancy. 

Third, while we confirm the existence of a statistically significant positional bias in document-level LLMs, its practical impact on translation quality appears to be negligible in current long-context models. 

Taken together, these findings suggest that the architectural shift towards LLMs alters the reliability of established quality prediction methods while also mitigating previous challenges in document-level MT.

\section{Limitations}
Our study's findings are subject to several limitations, primarily related to the dataset and the experimental scope.

All experiments are conducted on a single language pair, in a single translation direction (English-French) and a single, specific domain (encyclopedic content). The conclusions---particularly regarding the alignment of predictive models with NMT vs. LLMs and the negligible impact of positional bias---are robust for this dataset but may not generalize to other language pairs, other domains (e.g., creative, conversational, or legal texts), or other LLMs not included in our study.

Finally, a conceptual limitation lies in our operational definition of ``quality prediction'' itself. This study measures predictive power by correlating various heuristics and QE models against two automated, reference-based metrics: COMET and TER. These ground-truth metrics are, themselves, only proxies for the concepts they aim to capture and fallible (cf. \citealt{zouhar-etal-2024-pitfalls}, inter alia). A deep methodological critique of these foundational metrics is beyond the scope of this paper. We contend that our approach remains sound, as automated metrics are the \textit{de facto} standard in MT evaluation and the language service industry. Furthermore, our analysis benefits from presenting results against these two distinct and complementary definitions of quality. This dual approach prevents either metric from being taken as an absolute ``truth'' in isolation.

\section{Acknowledgements}
This work was funded by the French \textit{Agence Nationale de la Recherche} (ANR) under the project TraLaLaM (``ANR-23-IAS1-0006''), as well as by Inria under the ``\textit{Défi}''-type project COLaF. This work was partly funded by Rachel Bawden and Benoît Sagot’s chairs in the PRAIRIE institute, funded by the French national agency ANR, as part of the “Investissements d’avenir” programme under the reference ANR-19-P3IA-0001 and by Benoît Sagot's chair in its follow-up, PRAIRIE-PSAI, also funded by the ANR as part of the “France 2030” strategy under the reference ANR23-IACL-0008.

This work was also granted access to the HPC resources of IDRIS under allocations 2025-AD011015117R2 and 2025-AD011017228 made by GENCI.

\section{Bibliographical References}\label{sec:reference}

\bibliographystyle{lrec2026-natbib}
\bibliography{lrec2026-example}

\section{Language Resource References}
\label{lr:ref}
\bibliographystylelanguageresource{lrec2026-natbib}
\bibliographylanguageresource{languageresource}

\section{Appendix: Readability and Linguistic Complexity Metrics}
\label{sec:appendix_readability}

This appendix details the mathematical formulations of the standard readability metrics referenced in Section 3.1. All metrics were computed using the \texttt{textstat} Python library.\footnote{\href{https://github.com/textstat/textstat}{https://github.com/textstat/textstat}}

To present the formulas concisely, we define the following variables for a given text:
\begin{itemize}
    \setlength{\itemsep}{0pt}
    \item $W$: total number of words
    \item $S$: total number of sentences
    \item $Sy$: total number of syllables
\end{itemize}

\subsection*{Flesch Reading Ease (FRE)}
The FRE score (\textit{flesch\_ease} in the figures) evaluates readability based on the average number of syllables per word and words per sentence \cite{flesch1948new}. Higher scores indicate easier reading. Note that the constants used in this formula ($206.835$, $1.015$, and $84.6$) are specifically calibrated for the English language and defined by the library.
$$FRE = 206.835 - 1.015 \left(\frac{W}{S}\right) - 84.6 \left(\frac{Sy}{W}\right)$$

\subsection*{Flesch-Kincaid Grade Level (FKGL)}
The FKGL (\textit{fk\_grade} in the figures) translates the Flesch Reading Ease score into a U.S. grade-level equivalent \cite{kincaid1975derivation}. 
$$FKGL = 0.39 \left(\frac{W}{S}\right) + 11.8 \left(\frac{Sy}{W}\right) - 15.59$$

\subsection*{SMOG Index}
The Simple Measure of Gobbledygook (SMOG; \textit{smog\_index} in the figures) estimates the years of education needed to fully understand a piece of writing, relying heavily on the count of polysyllables ($P$), defined as words with three or more syllables \cite{mclaughlin1969smog}.
$$SMOG = 1.043 \sqrt{P \times \frac{30}{S}} + 3.1291$$

\subsection*{Coleman-Liau Index (CLI)}
Unlike Flesch and SMOG, the CLI (\textit{coleman\_liau} in the figures) relies on character counts instead of syllables, making it computationally robust for automated processing \cite{coleman1975computer}. Let $L$ be the average number of letters per 100 words, and $S_{100}$ be the average number of sentences per 100 words.
$$CLI = 0.0588L - 0.296S_{100} - 15.8$$

\subsection*{Dale-Chall Readability Formula}
This formula (represented as \textit{dale\_chall} in the figures) calculates a score based on average sentence length and the percentage of ``difficult'' words ($D$)---defined as words not present on a specific list of common words familiar to most fourth-grade students \cite{dale1948formula}.
$$\text{Score} = 0.1579 \left(\frac{D}{W} \times 100\right) + 0.0496 \left(\frac{W}{S}\right)$$
If the percentage of difficult words is greater than $5\%$, an adjusted score is calculated by adding $3.6365$ to the final result.

\subsection*{Linguistic Complexity Features}
In addition to the readability metrics, we calculated four linguistic complexity features referenced in Section 3.1. These were computed using a combination of the \texttt{spaCy},\footnote{\href{https://github.com/explosion/spaCy}{https://github.com/explosion/spaCy}} \texttt{NLTK} \citep{bird-loper-2004-nltk}, and \texttt{wordfreq} \citep{robyn_speer_2022_7199437} Python libraries. Let $W_{space}$ be the set of whitespace-separated tokens, and $W_{\alpha}$ be the subset of tokens containing only alphabetic characters.

\paragraph{Segment Length}
Segment length is defined strictly as the total number of whitespace-separated tokens in the source segment:
$$Length = |W_{space}|$$

\paragraph{Word Rarity}
Word rarity evaluates the average infrequency of the vocabulary used in the segment. It is calculated as the average negative natural logarithm of the word frequencies extracted via the \texttt{wordfreq} library. A smoothing factor of $10^{-9}$ is added to prevent undefined logarithms for out-of-vocabulary words.
$$Rarity = - \frac{1}{|W_{\alpha}|} \sum_{w \in W_{\alpha}} \ln(\text{freq}(w) + 10^{-9})$$

\paragraph{Degree of Polysemy}
The degree of polysemy captures lexical ambiguity by averaging the total number of WordNet synsets \citep{miller-1994-wordnet} available for each alphabetic token, retrieved using the \texttt{NLTK} library.
$$Polysemy = \frac{1}{|W_{\alpha}|} \sum_{w \in W_{\alpha}} |\text{synsets}(w)|$$

\paragraph{Syntactic Tree Height}
To measure structural complexity, we calculate the syntactic tree height. Using the \texttt{en\_core\_web\_lg} dependency parser from the \texttt{spaCy} library, we extract the dependency tree of the segment. The height is defined recursively as the maximum depth of the tree starting from the root node.

\section{Appendix: Supplementary Results using the ChrF Metric}
\label{sec:appendix_chrf}

The main body of this paper evaluates translation quality primarily through TER (as a proxy for post-editing effort) and COMET (as a proxy for human judgment). Due to space constraints in the main text, we present the supplementary results using the ChrF metric \citep{popovic-2015-chrf} in this appendix. 

The following figures replicate the primary correlation experiments from Sections 3 and 4, substituting the ground-truth reference metrics with ChrF to demonstrate how the predictive power of our source-side and candidate-side features behaves under this character-level surface-form evaluation paradigm. 

\begin{figure*}[t] 
\centering

\begin{subfigure}[b]{0.48\linewidth}
  \includegraphics[width=\linewidth]{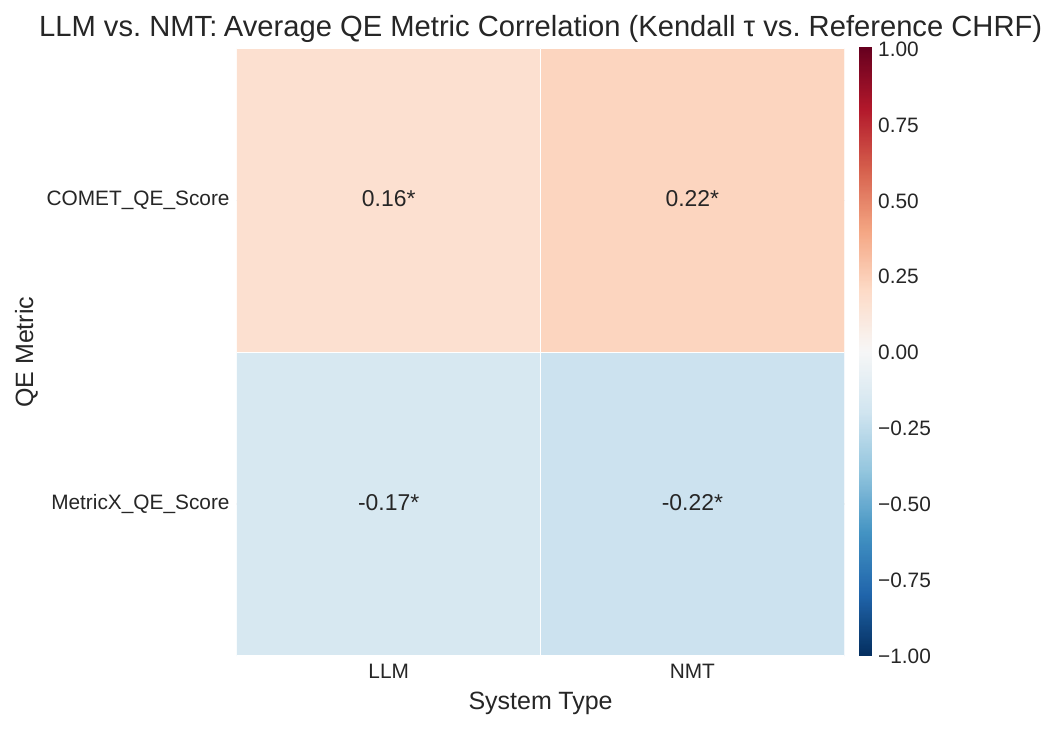}
  \caption{}
  \label{fig:sub4chrF}
\end{subfigure}
\hfill
\begin{subfigure}[b]{0.48\linewidth}
  \includegraphics[width=\linewidth]{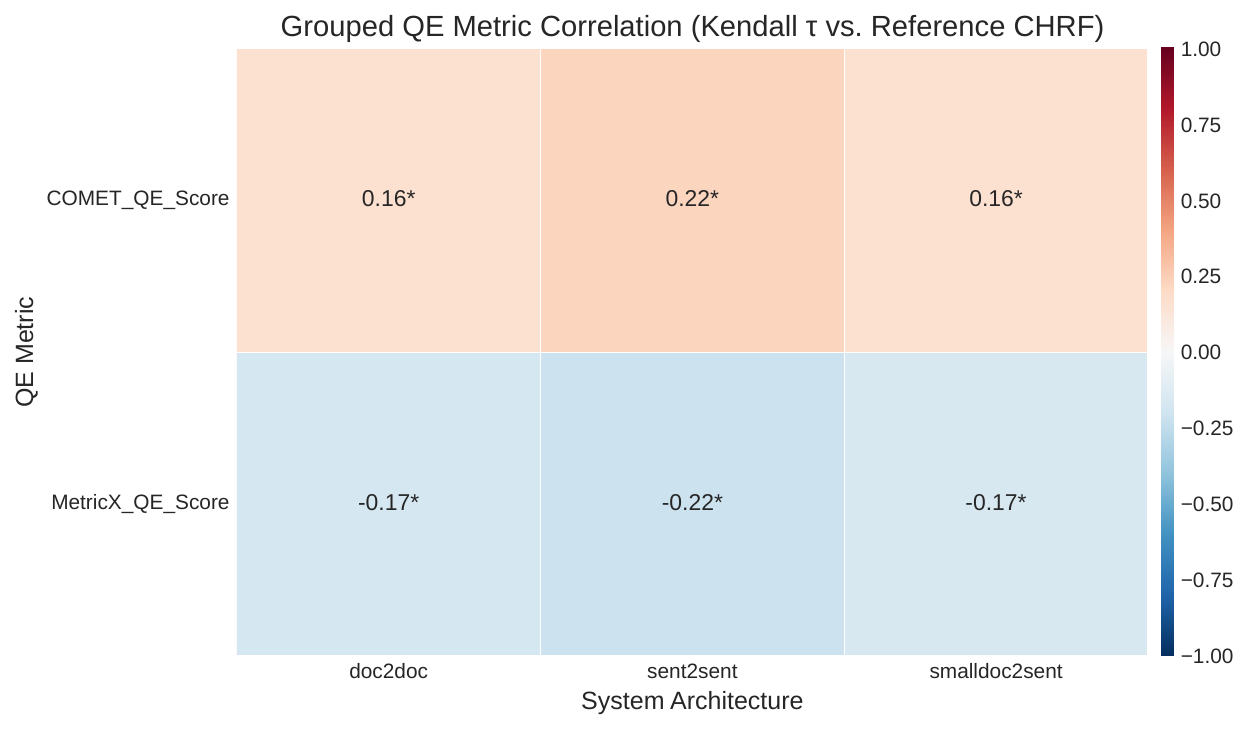}
  \caption{}
  \label{fig:sub5chrF}
\end{subfigure}

\caption{Kendall's $\tau$ correlation between QE metrics and gold-standard scores measured by ChrF, averaged across system groups. The left figure (a) groups by system type (LLM vs. NMT), while the right figure (b) groups by translation granularity.}
\label{fig:quadrantChrf}
\end{figure*}

\begin{figure*}
    \centering
    \includegraphics[width=1\linewidth]{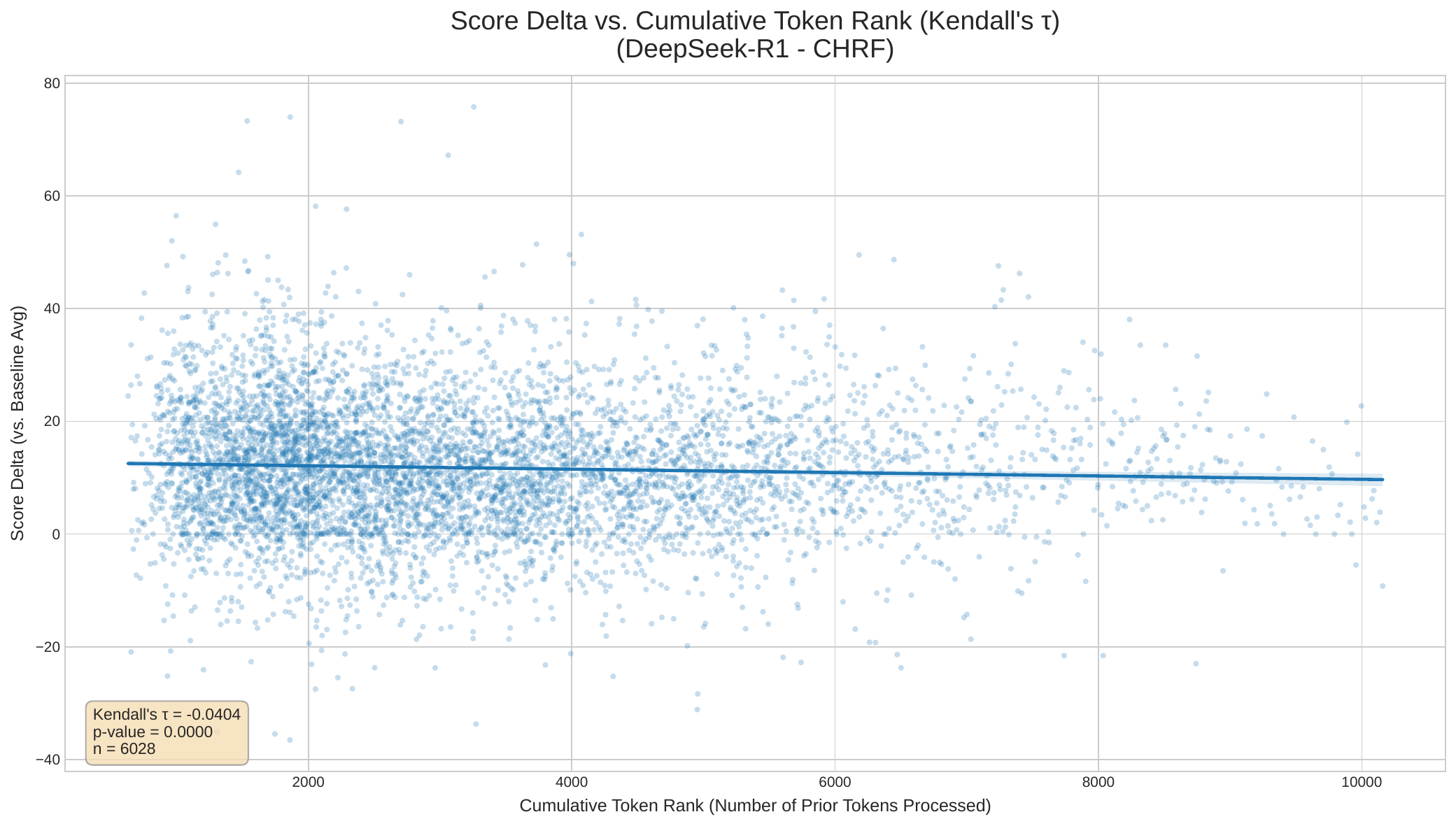}
    \caption{Scatter plots illustrating the effect of positional bias on the DeepSeek-R1 system. The plot shows the weak negative correlation between delta-normalized ChrF scores and the cumulative token rank, controlling for source-text difficulty. 
}
    \label{fig:chrf_posB}
\end{figure*}

\begin{figure*}
    \centering
    \includegraphics[width=1\linewidth]{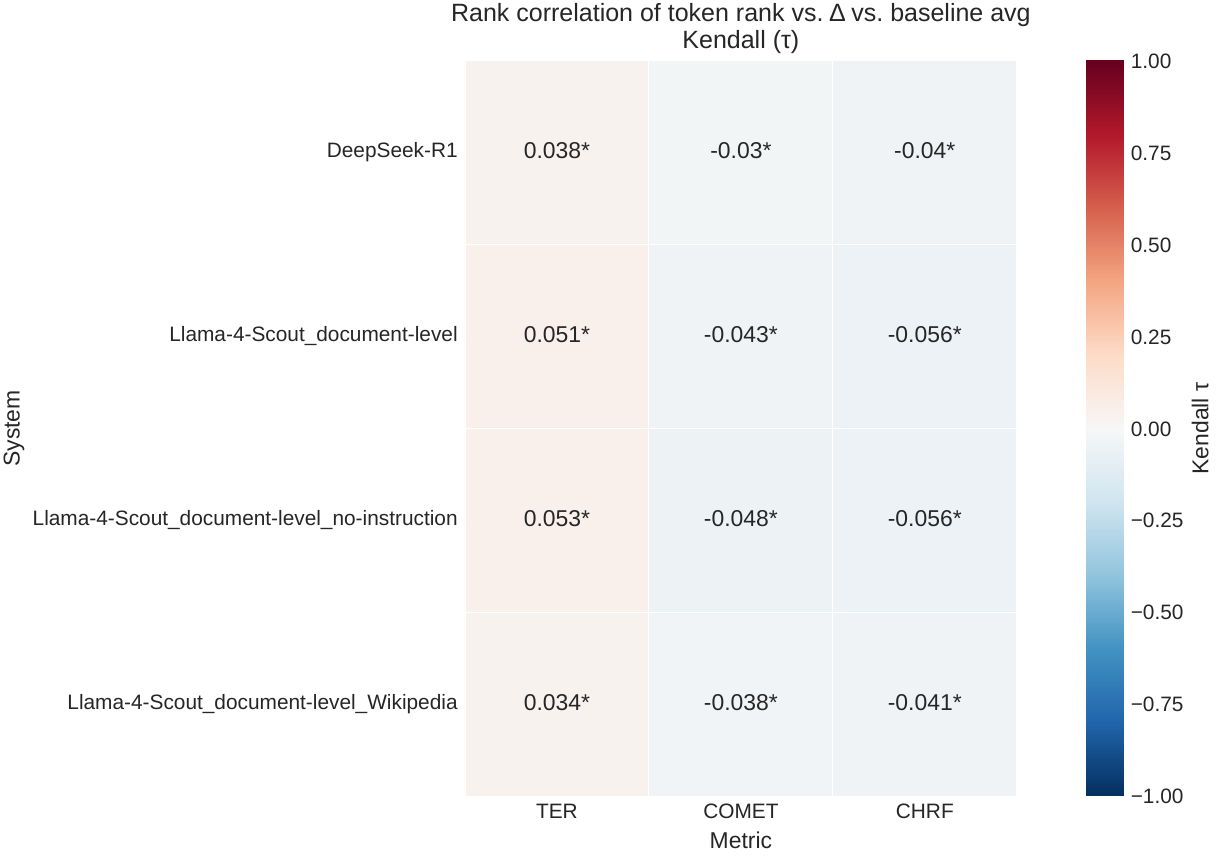}
    \caption{Kendall's $\tau$ correlation between a segment's cumulative token rank and its quality for all \texttt{doc2doc} systems. The plot uses our normalized delta scores and includes ChrF scores. 
}
    \label{fig:chrf_heatLB}
\end{figure*}

As illustrated in the supplementary Figures \ref{fig:chrf_aD}, \ref{fig:chrf_qeSys}, \ref{fig:quadrantChrf}, \ref{fig:chrf_posB} and \ref{fig:chrf_heatLB}, the predictive behavior of both source-side and candidate-side metrics against ChrF closely mirrors their behavior against TER, confirming the observations and metric dichotomies detailed in Sections 3 and 4.

\begin{figure*}
    \centering
    \includegraphics[width=1\linewidth]{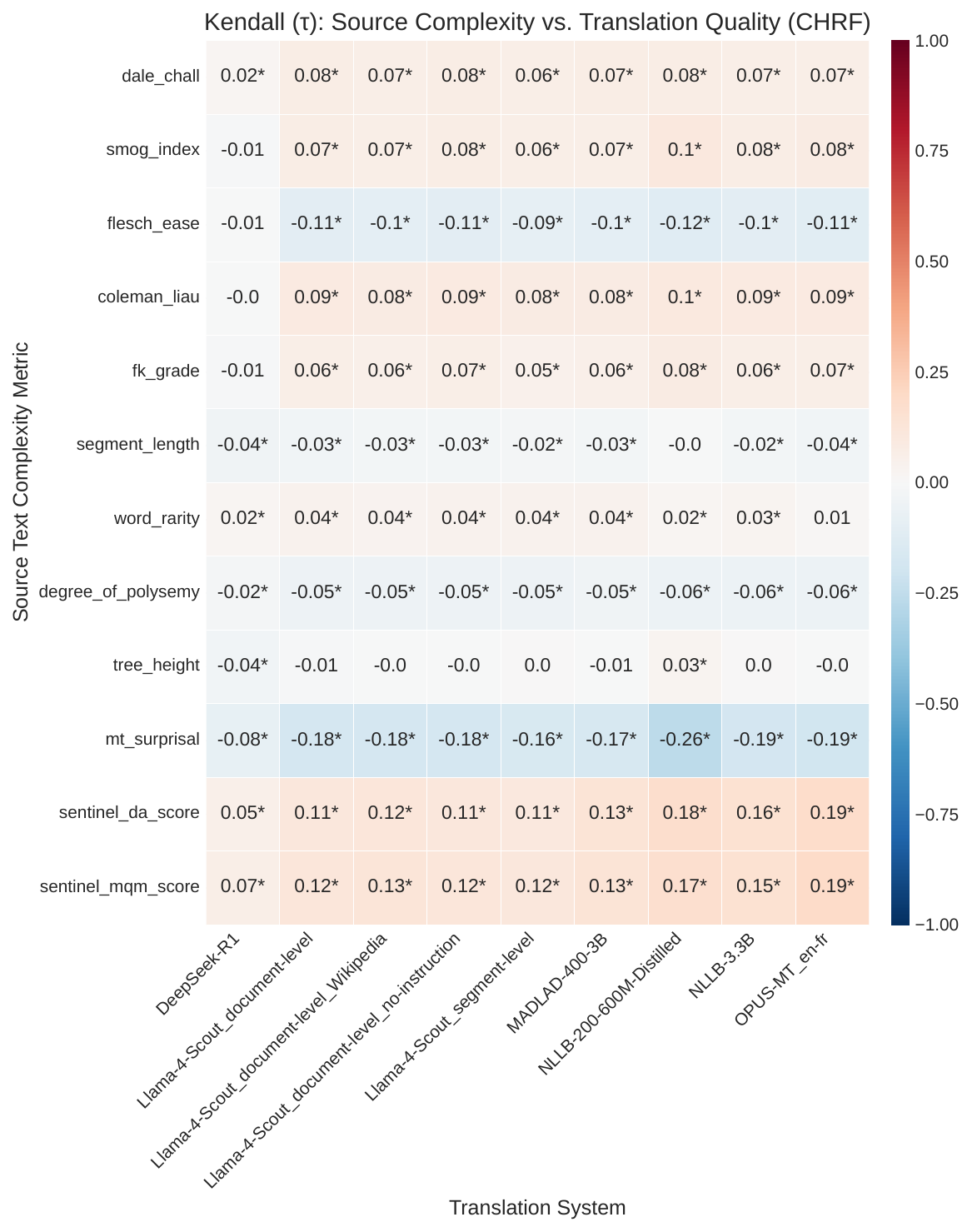}
    \caption{Kendall's~$\tau$ correlation between source-side metrics and translation quality as measured by ChrF. 
An asterisk~($^*$) indicates a statistically significant correlation ($p < 0.05$). 
}
    \label{fig:chrf_aD}
\end{figure*}

\begin{figure*}
    \centering
    \includegraphics[width=1\linewidth]{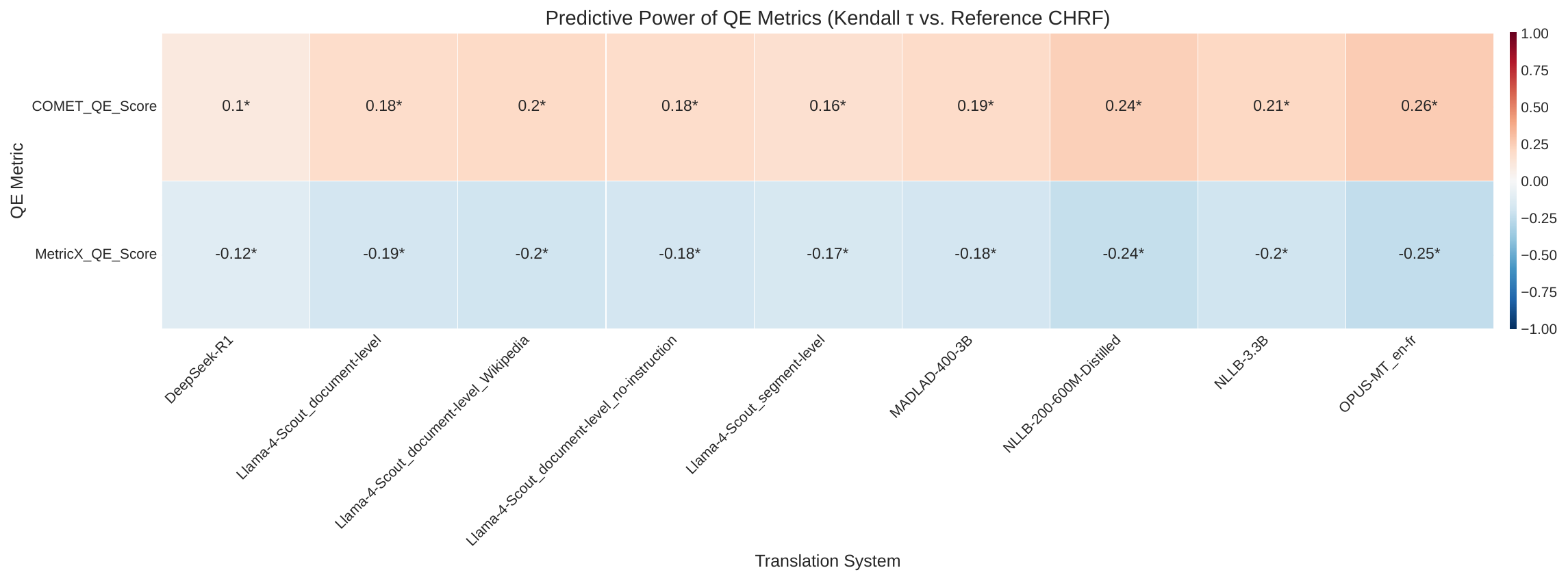}
    \caption{Kendall's $\tau$ correlation between two reference-free QE metrics (\textit{COMET\_QE}, \textit{MetricX\_QE}) and our gold-standard reference scores measured by ChrF. An asterisk (*) indicates statistical significance ($p<0.05$). 
}
    \label{fig:chrf_qeSys}
\end{figure*}

\end{document}